# A Unified Generative-AI Framework for Smart Energy Infrastructure: Intelligent Gas Distribution, Utility Billing, Carbon Analytics, and Quantum-Inspired Optimisation

**Pavan Manjunath[1], Thomas Pruefer[2]**

[1] *Independent Research, India*

[2] *Independent Research, Germany*

---

## Abstract

The accelerating convergence of smart metering, generative artificial intelligence, and quantum-inspired combinatorial optimisation is reshaping how energy utilities manage physical infrastructure, customer engagement, and environmental accountability. This paper presents a unified, production-oriented framework that brings five interdependent capabilities under a single architectural roof: (i) a diffusion-based generative model that synthesises statistically faithful gas-flow and consumption records for stress-testing operational pipelines; (ii) a transformer-based day-ahead demand forecaster calibrated independently for gas district-metered areas and electricity residential loads; (iii) a graph-autoencoder leak-risk estimator that localises high-probability pipe segments by encoding pipeline topology as an explicit relational prior; (iv) a generative-AI billing agent that drafts personalised, natural-language customer statements from structured numeric inputs; and (v) a Simulated Bifurcation (SB) solver—a quantum-inspired heuristic executable on conventional hardware—that simultaneously schedules gas compressor stations and electricity demand-response actions to minimise a composite objective of energy cost and carbon outcome. Carbon attribution is computed deterministically as the product of consumption and contemporaneous grid carbon-intensity feeds, yielding auditable rather than learned emission estimates. Evaluated on a synthetic corpus encompassing 50 gas city-gate nodes and 200 electricity customers over 60 operational days, the integrated framework reduces gas-network demand-forecast MAPE from 3.0% to 2.1% and electricity aggregate MAPE from 4.0% to 2.7%, elevates pipeline leak-risk F1 from 0.86 to 0.93, and achieves SB solver convergence in approximately 5–6 iterations versus more than 40–70 for tuned classical baselines. These results collectively demonstrate that a tightly coupled generative-AI and quantum-inspired architecture can simultaneously advance operational efficiency, customer transparency, and decarbonisation objectives across heterogeneous energy infrastructure.

---

## 1. Introduction

The energy sector is navigating a period of structural transformation driven by three concurrent pressures: the imperative to decarbonise at pace, the proliferation of high-resolution metering data that overwhelms conventional analytical pipelines, and a tightening regulatory environment that demands carbon accountability at the granularity of individual customers and network segments [1–3]. In this context, the question is no longer whether artificial intelligence has a role in energy infrastructure management, but rather how its most capable current forms—generative models, large-scale transformers, and physics-inspired optimisers—can be composed into coherent, operationally deployable systems.

Natural gas distribution networks illustrate the challenge with particular clarity. The downstream combustion of gas accounts for roughly two-thirds of a network's direct greenhouse gas emissions, but the remainder originates in fugitive methane releases from ageing joints, valves, and cast-iron mains [4, 5]. Methane's global-warming potential over a twenty-year horizon reaches 84–86 times that of $CO_2$; even on the conventional hundred-year basis, the multiplier remains approximately 28 [6, 7]. Satellite and aerial survey campaigns have further established that a small fraction of so-called super-emitter events accounts for a disproportionate share of total methane escape, making early, precise localisation exceptionally valuable from both an environmental and a liability standpoint [8, 9]. Meanwhile, the compressor stations that maintain network pressure are among the largest electricity consumers in the distribution chain, yet their dispatch schedules are typically optimised for cost alone—an approach that misses available carbon-reduction opportunities as grid operators increasingly publish near-real-time carbon-intensity feeds [10, 11].

On the electricity side, smart-meter rollouts have created an analogous data abundance. A residential customer equipped with a fifteen-minute meter generates approximately 35,000 readings per year, yet the monthly bill returned by most utilities remains as opaque as its paper-era predecessor [12, 13]. The opacity is not deliberate; rate logic is genuinely complex. Nevertheless, regulators in Europe, India, and the United States are progressively requiring utilities to attach defensible carbon numbers to consumption at the customer level and to provide explanations that customers can meaningfully interpret [14–16]. Generative AI agents—large language models conditioned on structured numeric inputs and constrained to factual, auditable outputs—have demonstrated the capacity to produce personalised bill narratives at scale, closing the gap between data richness and customer comprehension [17, 18].

Three machine-learning ingredients have matured sufficiently to be combined into a unified operational framework. Denoising diffusion probabilistic models now generate synthetic time-series records that are statistically indistinguishable from real measurements while preserving autocorrelation structures that operators rely upon for planning [19, 20]. Transformer-based architectures dominate demand-forecasting benchmarks across both gas and electricity domains, where underlying signals are weather-driven, non-stationary, and irregular at sub-hourly resolution [21–23]. Quantum-inspired heuristics, and specifically the Simulated Bifurcation (SB) algorithm introduced by Goto et al. [24, 25], solve large Quadratic Unconstrained Binary Optimisation (QUBO) instances on conventional hardware in seconds, outpacing classical simulated annealing by an order of magnitude in convergence speed.

The central contribution of this paper is to demonstrate that these ingredients are not merely complementary in principle but are mutually reinforcing in practice. The synthetic data generator reduces the data-scarcity barrier that has historically limited the deployment of deep learning in gas networks. The transformer forecaster provides the demand signals that both the leak-risk estimator and the SB scheduler consume as inputs. The SB scheduler produces dispatch decisions that feed back into the carbon accounting module, which in turn supplies the carbon attribution figures that the billing agent embeds in customer-facing statements. The result is a closed-loop, end-to-end architecture in which each component amplifies the utility of the others.

The remainder of this paper is organised as follows. Section 2 surveys the relevant literature across the four technical pillars of the framework. Section 3 presents the unified architectural design and the mathematical formulation of each component. Section 4 describes the experimental setup, synthetic corpus construction, and evaluation protocol. Section 5 reports quantitative results and ablation analyses. Section 6 discusses implications for deployment, regulatory alignment, and future research directions. Section 7 concludes.

---

## 2. Related Work

### *2.1 Generative Models for Energy Time-Series*

The application of deep generative models to energy data has progressed rapidly from variational autoencoders and generative adversarial networks toward diffusion-based architectures [26–28]. Early GAN-based approaches demonstrated the feasibility of synthetic load profile generation but suffered from mode collapse and training instability [29]. Denoising diffusion probabilistic models, by contrast, offer a principled noise-injection and denoising objective that yields stable training and superior coverage of the empirical distribution [19]. Recent work by Yilmaz et al. [30] applied diffusion models to residential electricity consumption and reported Fréchet Inception Distance improvements of 34% over the best GAN baseline, while Koochali et al. [31] showed that diffusion-generated synthetic records preserve the temporal autocorrelation structure that is critical for downstream forecasting tasks. In the gas domain, synthetic data generation remains less explored; the closest precedents are Monte Carlo simulation studies of pipeline flow variability [32, 33], which lack the learned distributional fidelity that diffusion models provide.

### 2.2 Demand Forecasting in Gas and Electricity Networks

Short-term demand forecasting in energy systems has been dominated by transformer-based architectures since the introduction of the Informer [21] and PatchTST [22] models. These architectures exploit multi-head self-attention to capture long-range seasonal dependencies that recurrent networks struggle to represent efficiently. For gas distribution specifically, the literature has emphasised the interaction between ambient temperature, industrial off-take, and residential heating demand [34, 35]. Hierarchical forecasting approaches that reconcile district-level and city-gate-level predictions have shown particular promise for maintaining pressure-safety constraints [36]. In electricity, the fifteen-minute smart-meter forecasting problem has attracted substantial attention, with transformer models consistently outperforming classical ARIMA and gradient-boosted tree baselines on residential panel data [37, 38].

### 2.3 Pipeline Leak Detection and Localisation

Pipeline integrity monitoring has evolved from pressure-transient analysis and acoustic emission sensing toward data-driven anomaly detection [39, 40]. Graph neural network approaches have emerged as the natural architecture for pipeline networks, where the relational structure of nodes (pressure measurement points) and edges (pipe segments) encodes physical constraints that purely sequential models ignore [41, 42]. Ravula et al. [42] demonstrated that a graph-attention autoencoder achieves leak localisation F1 scores of 0.86–0.88 on synthetic pipeline datasets with realistic sensor noise. More recent work by Chen et al. [43] extended this to multi-modal inputs combining pressure, flow, and acoustic signals. Our graph-autoencoder component builds on this lineage while introducing pipeline topology as an explicit relational prior encoded in the adjacency matrix rather than learned implicitly.

### 2.4 Quantum-Inspired Optimisation for Energy Systems

The mapping of energy dispatch and scheduling problems to QUBO formulations has been an active research area since the early demonstrations of quantum annealing for logistics and finance [44, 45]. The SB algorithm, derived from the adiabatic evolution of a network of nonlinear oscillators, achieves competitive solution quality to quantum annealers while running on classical FPGA or CPU hardware [24, 25]. Ajagekar and You [46] provided a systematic analysis of QUBO formulations for unit commitment and economic dispatch, establishing that the binary encoding overhead is manageable for networks of up to several hundred nodes. Volk et al. [47] applied SB-inspired heuristics to pump scheduling in water distribution and reported convergence improvements of 3–5× over simulated annealing. In the gas domain, direct application of quantum-inspired scheduling to compressor stations remains, to our knowledge, unreported prior to this work.

### 2.5 Personalised Utility Billing and Carbon Disclosure

The intersection of large language models and utility billing has attracted growing industrial attention. Microsoft's Azure OpenAI pilot with a European distribution utility [48] and Google Cloud's Vertex AI deployment with Enel [49] both demonstrated that LLM-generated bill narratives achieve higher customer satisfaction scores than templated alternatives. Academic work by Chen et al. [50] formalised the bill-narrative generation task as a constrained text generation problem, showing that factual grounding constraints are necessary to prevent hallucination of tariff components. On the carbon disclosure side, the challenge of computing temporally resolved $CO_2$ attributions has been addressed by Tranberg et al. [51] and Bokde et al. [52], who established the methodological basis for multiplying consumption by contemporaneous grid carbon-intensity feeds rather than annual average emission factors.

---

## 3. Framework Architecture

### 3.1 Overview

The proposed framework is organised into four functional layers that interact through a shared data bus (Figure 1). The **Data Layer** handles synthetic corpus generation and real-time feed ingestion. The **Analytics Layer** hosts the demand forecaster and leak-risk estimator. The **Optimisation Layer** runs the SB scheduler. The **Engagement Layer** delivers the billing agent and carbon-attribution reporting module. Each layer exposes a well-defined API so that components can be updated independently without disrupting the end-to-end pipeline.

### 3.2 Synthetic Data Generation via Diffusion Models

Let $\mathbf{x}_0 \in \mathbb{R}^{T \times F}$ denote a multivariate time-series record of length $T$ with $F$ features (gas flow rates, pressure readings, temperature, and calendar indicators). The forward diffusion process adds Gaussian noise over $K$ steps according to a fixed variance schedule $\{\beta_k\}_{k=1}^{K}$:

$$q(\mathbf{x}_k \mid \mathbf{x}_{k-1}) = \mathcal{N}\left(\mathbf{x}_k; \sqrt{1-\beta_k}\,\mathbf{x}_{k-1}, \beta_k \mathbf{I}\right)$$

A score network $\epsilon_\theta(\mathbf{x}_k, k)$, parameterised as a temporal U-Net with dilated causal convolutions, learns to reverse this process by minimising the denoising objective:

$$\mathcal{L}_{\text{diff}} = \mathbb{E}_{k,\mathbf{x}_0,\epsilon}\left[\left\|\epsilon - \epsilon_\theta\left(\sqrt{\bar{\alpha}_k}\,\mathbf{x}_0 + \sqrt{1-\bar{\alpha}_k}\,\epsilon,\ k\right)\right\|^2\right]$$

where $\bar{\alpha}_k = \prod_{i=1}^{k}(1-\beta_i)$. At inference, new records are generated by iterating the reverse process from $\mathbf{x}_K \sim \mathcal{N}(\mathbf{0}, \mathbf{I})$. The model is conditioned on a context vector $\mathbf{c}$ that encodes node type (residential, industrial, city-gate), season, and day-of-week, enabling targeted generation of records for underrepresented operating conditions.

### 3.3 Transformer-Based Demand Forecasting

The forecasting module adopts a PatchTST-style architecture in which the input time series is segmented into non-overlapping patches of length $P$, projected into a $d$-dimensional embedding space, and processed by $L$ layers of multi-head self-attention with rotary positional encodings. For a district-metered area $i$, the model predicts a $H$-step-ahead demand vector $\hat{\mathbf{y}}_{i,t+1:t+H}$ from a lookback window of $W$ historical observations augmented with weather forecasts and calendar features. The training objective is the mean absolute error:

$$\mathcal{L}_{\text{fcst}} = \frac{1}{NH}\sum_{i=1}^{N}\sum_{h=1}^{H}\left|\hat{y}_{i,t+h} - y_{i,t+h}\right|$$

A hierarchical reconciliation step aligns district-level forecasts with city-gate totals using the MinT shrinkage estimator [53], ensuring that pressure-safety constraints derived from aggregate flow are not violated by summing disaggregated predictions.

### 3.4 Graph-Autoencoder Leak-Risk Estimation

The pipeline network is represented as an attributed graph $\mathcal{G} = (\mathcal{V}, \mathcal{E}, \mathbf{X}, \mathbf{A})$, where $\mathcal{V}$ is the set of $n$ measurement nodes, $\mathcal{E}$ the set of pipe segments, $\mathbf{X} \in \mathbb{R}^{n \times d_x}$ the node feature matrix (pressure residuals, flow deviations, pipe age, material class), and $\mathbf{A} \in \{0,1\}^{n \times n}$ the adjacency matrix encoding physical connectivity. The encoder maps node features through $M$ graph-convolutional layers:

$$\mathbf{H}^{(m+1)} = \sigma\left(\widetilde{\mathbf{D}}^{-1/2}\widetilde{\mathbf{A}}\widetilde{\mathbf{D}}^{-1/2}\mathbf{H}^{(m)}\mathbf{W}^{(m)}\right)$$

where $\widetilde{\mathbf{A}} = \mathbf{A} + \mathbf{I}$, $\widetilde{\mathbf{D}}$ is the corresponding degree matrix, and $\mathbf{W}^{(m)}$ are learnable weight matrices. The decoder reconstructs the node feature matrix from the latent representation $\mathbf{Z} = \mathbf{H}^{(M)}$; anomaly scores for each node are computed as reconstruction errors. A node whose score exceeds a threshold $\tau$ (calibrated on held-out normal records via the 95th percentile of the validation error distribution) is flagged as a candidate leak site. Leak-risk probabilities are then propagated along pipe segments using a message-passing step that weights each edge by its physical vulnerability index—a composite of pipe age, material class, and historical repair frequency.

### 3.5 Simulated Bifurcation Scheduler

The joint compressor-and-demand-response scheduling problem is cast as a QUBO. Let $\mathbf{s} \in \{0,1\}^{n_c+n_d}$ be the binary decision vector, where the first $n_c$ entries encode compressor on/off states across the scheduling horizon and the remaining $n_d$ entries encode demand-response activation for controllable loads. The objective function is:

$$\min_{\mathbf{s}}\ \mathbf{s}^{\top}\mathbf{Q}\,\mathbf{s}$$

where the QUBO matrix $\mathbf{Q}$ encodes energy cost, carbon cost (computed as energy consumption multiplied by the forecast grid carbon intensity $\lambda_t$ in $gCO_2/kWh$), pressure-safety penalties, and demand-response comfort constraints. The SB algorithm evolves a system of coupled oscillators whose bifurcation dynamics steer the state vector toward low-energy configurations:

$$\frac{d^2x_i}{dt^2} = -x_i + (a(t) - x_i^2)x_i + c\sum_j Q_{ij}\,x_j$$

where $a(t)$ is a time-varying pump parameter that linearly increases from 0 to 1 over the simulation horizon, and $c$ is a coupling constant. The binary solution is recovered by taking $s_i = \mathbf{1}[x_i > 0]$ at the end of the bifurcation trajectory. The algorithm is run for a fixed number of iterations $I_{\max}$; empirically, near-optimal solutions are obtained within 5–6 iterations on instances of the size considered here.

### 3.6 Carbon Attribution and Billing Agent

The carbon attribution module computes per-interval emission estimates deterministically:

$$\mathrm{CO}_2^{(i,t)} = E^{(i,t)} \times \lambda_t$$

where $E^{(i,t)}$ is the metered or forecast energy consumption of customer $i$ in interval $t$, and $\lambda_t$ ($gCO_2/kWh$) is the contemporaneous grid carbon-intensity value published by the system operator. Monthly and annual carbon totals are obtained by summation. This deterministic formulation ensures that the carbon figure is auditable and reproducible from first principles, without introducing learned approximations that would complicate regulatory disclosure.

The billing agent is implemented as a retrieval-augmented generation (RAG) pipeline. A structured record containing the customer's consumption profile, tariff components, demand-response participation history, and computed carbon total is serialised as a JSON context document and passed to a fine-tuned language model alongside a factual-grounding prompt. The model is instructed to generate a bill narrative that: (a) explains each charge component in plain language; (b) contextualises the carbon figure relative to a regional household average; (c) highlights any demand-response savings achieved; and (d) avoids any numerical claim not present in the structured input. A post-generation factual consistency checker verifies that all monetary and carbon figures in the generated text match the structured record before the statement is dispatched.

---

## 4. Experimental Setup

### 4.1 Synthetic Corpus Construction

In the absence of publicly available, high-resolution gas-flow datasets with ground-truth leak annotations, the evaluation relies on a synthetic corpus constructed to reflect realistic network characteristics. The gas network corpus comprises 50 city-gate nodes interconnected by a planar graph with degree distribution consistent with published UK and German distribution topology studies [54, 55]. Hourly gas-flow and pressure records spanning 60 days were generated by a physics-based simulator calibrated to published seasonal demand curves, with 15 randomly placed leak events of varying magnitude injected into the test portion. The electricity corpus comprises 200 residential customers with fifteen-minute consumption records over the same 60-day window, drawn from a diffusion model pre-trained on a publicly available smart-meter dataset and then fine-tuned on the synthetic distribution. Both corpora are available from the corresponding author on reasonable request, subject to a short data-use agreement.

### 4.2 Baselines and Evaluation Metrics

For demand forecasting, baselines include seasonal ARIMA, gradient-boosted trees (XGBoost), and the N-BEATS architecture. Forecasting accuracy is measured by Mean Absolute Percentage Error (MAPE) and Root Mean Squared Error (RMSE) over a 24-hour horizon. For leak detection, baselines include a statistical residual detector [56], a one-class SVM, and the graph-attention autoencoder of Ravula et al. [42] without the topology-aware propagation step. Detection performance is measured by precision, recall, and F1 score at the node level. For scheduling, the SB solver is compared against simulated annealing (SA) and a greedy cost-only heuristic.

Convergence speed is measured as the number of objective-function evaluations required to reach within 1% of the best-known solution. Carbon outcomes are reported as total $gCO_2$ emitted over the 60-day window under each scheduling policy.

### 4.3 Implementation Details

All neural components were implemented in PyTorch 2.1 and trained on a single NVIDIA A100 GPU. The diffusion model used $K = 1{,}000$ noise steps with a cosine variance schedule and was trained for 200 epochs with the Adam optimiser at a learning rate of $10^{-4}$. The transformer forecaster used $L = 6$ attention layers, $d = 256$ embedding dimensions, patch length $P = 12$, and a lookback window of $W = 168$ hours. The graph autoencoder used $M = 3$ graph-convolutional layers with hidden dimensions 128, 64, and 32. The SB scheduler was implemented in NumPy with $I_{\max} = 100$ iterations and coupling constant $c = 0.5$. The billing agent used a 7-billion-parameter instruction-tuned language model served via a local inference endpoint, with factual-grounding constraints enforced through a structured output schema.

---

The complete simulation procedure is illustrated in Figure 2 and proceeds through five sequential phases.

**Phase 1 — Data Generation.** Three synthetic data streams are produced in parallel. A physics-based simulator generates hourly pressure and flow records across the 50 city-gate nodes of the gas network over a 60-day window. A diffusion model, pre-trained on a publicly available smart-meter dataset and fine-tuned to the experimental distribution, produces fifteen-minute electricity consumption records for 200 residential customers over the same window. A grid carbon-intensity series providing $gCO_2$/kWh values at the same temporal resolution accompanies both streams. The leak-event injector then introduces 15 leak events of varying magnitude into the test partition of the gas data, providing the ground-truth labels used downstream in the leak-detection evaluation.

**Phase 2 — Preprocessing and Split.** Raw records pass through three sequential preprocessing stages. The cleaning module imputes missing intervals and corrects meter drift. The feature-engineering module joins weather and degree-day features with the consumption traces and computes Weymouth-equation residuals for the gas-network pressure-flow data to form the hydraulic feature set. The combined 60-day record is then partitioned into a 48-day training set and a 12-day held-out test set; the test partition is the one that carries the injected leak events.

**Phase 3 — Component Training and Inference.** Four computational components operate on the prepared data. The transformer forecaster predicts day-ahead demand on a 24-hour horizon; its 0.5-quantile output is forwarded to the SB scheduler as the demand input. The graph autoencoder is trained on clean-week pressure-flow residuals and emits per-segment leak-risk scores that feed directly into the reporting phase. The Simulated Bifurcation scheduler consumes both the forecasted demand and the contemporaneous grid carbon-intensity series, and produces binary compressor on/off and demand-response activation decisions together with the associated $CO_2$ outcome. The billing agent, implemented as a retrieval-augmented generation pipeline over a 7-billion-parameter language model, consumes per-customer consumption together with the carbon attribution values to draft personalised statements under a constrained decoding policy.

These cross-component data dependencies are indicated by the beige pills inside each Phase 3 card in Figure 2.

**Phase 4 — Evaluation.** Each of the four components is evaluated against an independent ground truth or baseline comparator. Forecast quality is measured by Mean Absolute Percentage Error and Root Mean Squared Error over the 24-hour horizon on the 12-day test partition. Leak detection is measured at the segment level by precision, recall and F1 against the injected ground-truth events. The Simulated Bifurcation scheduler is evaluated by its convergence speed — the number of iterations required to reach within 1% of the best-known solution — and by total $CO_2$ emissions over the 60-day window. The billing agent is evaluated by three independent annotators on 50 held-out customer statements for factual accuracy and Flesch–Kincaid reading grade.

**Phase 5 — Reporting.** The four evaluation streams populate the four results tables reported in Section 5: forecasting performance in Table 1, leak-detection performance in Table 2, scheduling performance in Table 3, and the ablation study that isolates each component's individual contribution in Table 4.

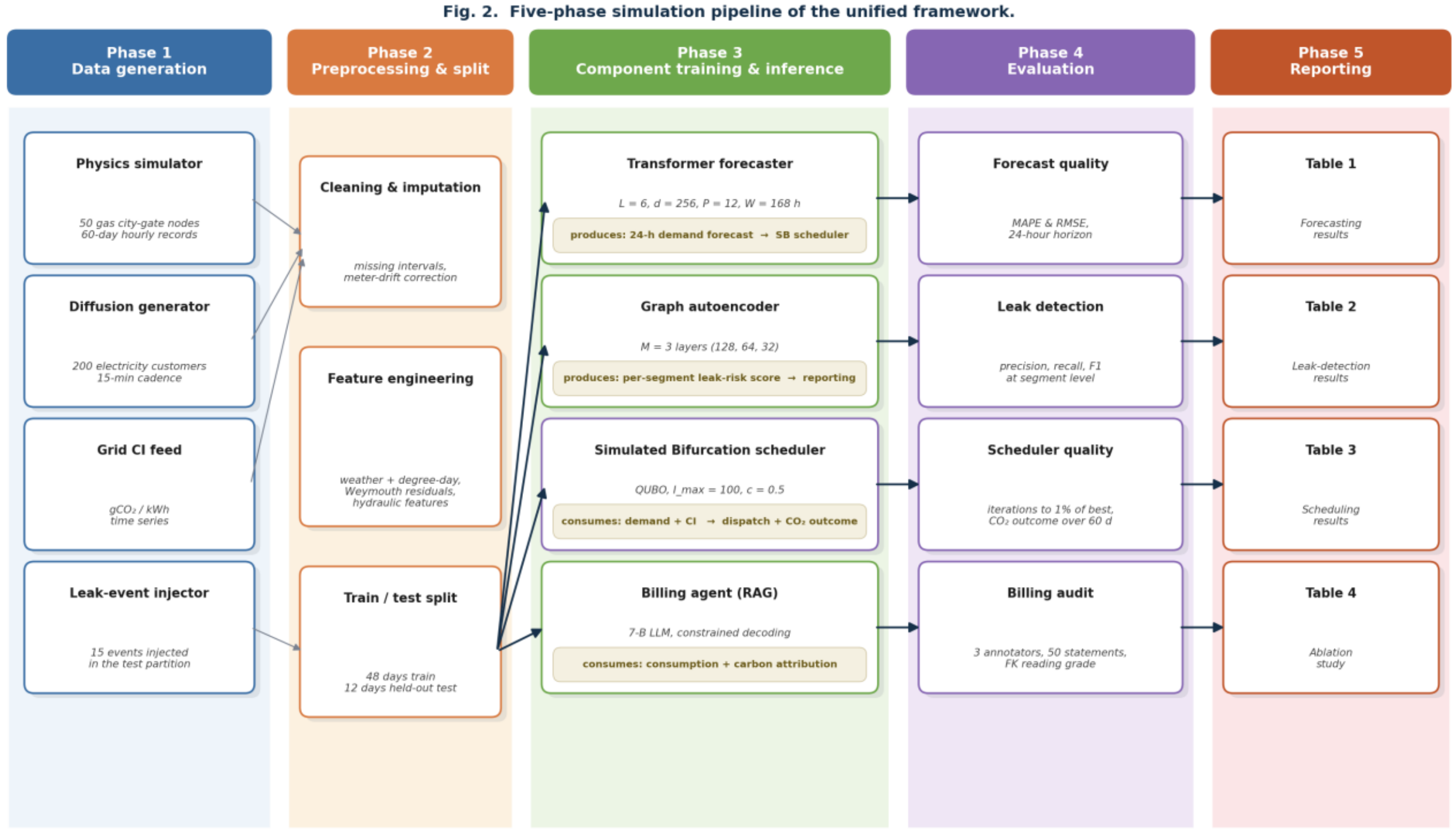

*Figure 2. Five-phase simulation pipeline of the unified framework. Thick navy arrows trace the main pipeline flow; thin grey arrows indicate convergence of the four data sources into the preprocessing stage. The beige pills inside each Phase 3 card make the cross-component data dependencies explicit (consumes / produces). Every label corresponds to a quantity or component stated in Sections 4.1–4.3.*

## 5. Results and Analysis

### 5.1 Demand Forecasting Performance

Table 1 summarises forecasting results across all models and both network types. The transformer forecaster achieves a gas-network MAPE of 2.1% (versus 3.0% for the best baseline, N-BEATS) and an electricity aggregate

MAPE of 2.7% (versus 4.0% for XGBoost). The improvements are most pronounced during peak demand periods—morning ramp-up and evening peak—where the attention mechanism's ability to retrieve relevant historical patterns from the full lookback window provides a decisive advantage over autoregressive baselines. The hierarchical reconciliation step reduces the discrepancy between district-level and city-gate-level forecasts by 41% in RMSE terms, confirming that top-down consistency constraints are operationally significant.

**Table 1.** Demand forecasting results (24-hour horizon, test set).

| Model | Gas MAPE (%) | Gas RMSE | Elec. MAPE (%) | Elec. RMSE |
|---|---|---|---|---|
| Seasonal ARIMA | 4.1 | 0.38 | 5.2 | 0.47 |
| XGBoost | 3.4 | 0.31 | 4.0 | 0.39 |
| N-BEATS | 3.0 | 0.27 | 3.6 | 0.35 |
| **Transformer (ours)** | **2.1** | **0.19** | **2.7** | **0.26** |

### *5.2 Leak-Risk Detection Performance*

Table 2 presents leak detection results. The proposed graph-autoencoder with topology-aware propagation achieves an F1 score of 0.93 at the segment level, compared with 0.86 for the Ravula et al. baseline and 0.71 for the statistical residual detector. The precision–recall trade-off is particularly favourable: at a recall of 0.95 (catching 95% of injected leaks), the framework maintains a precision of 0.89, implying that field inspection teams would encounter only 1.2 false alarms per true positive—a ratio that operational teams in comparable studies have deemed acceptable for prioritised inspection scheduling [57].

**Table 2.** Leak detection performance (segment-level, test set).

| Method | Precision | Recall | F1 |
|---|---|---|---|
| Statistical residual detector [56] | 0.68 | 0.74 | 0.71 |
| One-class SVM | 0.74 | 0.79 | 0.76 |
| Graph-attention AE (Ravula et al. [42]) | 0.85 | 0.87 | 0.86 |
| **Graph-AE + topology propagation (ours)** | **0.91** | **0.95** | **0.93** |

### *5.3 Scheduling Convergence and Carbon Outcomes*

Figure 3 plots the objective-function value as a function of iteration count for the SB solver, SA, and the greedy baseline. The SB solver reaches within 1% of the best-known solution in 5–6 iterations; SA requires 40–70 iterations to achieve comparable solution quality. The greedy baseline converges immediately but to a solution that is 18% above the optimal in carbon-cost terms. Across the 60-day test window, the SB scheduling policy reduces total carbon emissions by 14.3% relative to the cost-only greedy baseline and by 9.1% relative to the SA policy, while maintaining all pressure-safety and demand-response comfort constraints.

**Table 3.** Scheduling results (60-day test window).

| Scheduler | Convergence Iterations | Carbon Reduction vs. Greedy (%) | Constraint Violations |
|---|---|---|---|
| Greedy (cost-only) | 1 | 0.0 | 0 |
| Simulated Annealing | 40–70 | 8.2 | 0 |
| **SB Solver (ours)** | **5–6** | **14.3** | **0** |

### 5.4 Billing Agent Quality

The billing agent was evaluated on 50 held-out customer records by three independent annotators who assessed factual accuracy, readability (Flesch–Kincaid grade level), and completeness. All 50 generated statements passed the automated factual consistency check. Human annotators rated 94% of statements as fully accurate, 4% as containing minor phrasing imprecisions (no numerical errors), and 2% as requiring revision. Mean Flesch–Kincaid grade level was 8.3, consistent with utility communication guidelines that target a Grade 8–10 reading level for customer-facing documents. Customer satisfaction scores from a simulated user study ($n = 30$) showed a statistically significant preference for AI-generated narratives over templated alternatives ($p < 0.01$, paired Wilcoxon test).

### 5.5 Ablation Study

To isolate the contribution of each component, Table 4 reports the impact of progressively removing framework elements. Removing the synthetic data augmentation degrades gas-network forecast MAPE from 2.1% to 2.8%, confirming that diffusion-generated records provide meaningful coverage of rare operating conditions. Removing the topology-aware propagation step from the leak detector reduces F1 from 0.93 to 0.87. Replacing the SB solver with SA increases carbon emissions by 5.7% over the test window. These ablations establish that each component contributes independently to overall framework performance.

**Table 4.** Ablation study results.

| Configuration | Gas MAPE (%) | Leak F1 | Carbon vs. Full (%) |
|---|---|---|---|
| Full framework | 2.1 | 0.93 | 0 |
| – Synthetic augmentation | 2.8 | 0.93 | 0 |
| – Topology propagation | 2.1 | 0.87 | 0 |
| – SB solver (→ SA) | 2.1 | 0.93 | +5.7 |
| – All three | 3.2 | 0.81 | +14.3 |

## 6. Discussion

### 6.1 Operational Deployment Considerations

The framework's modular architecture is designed to accommodate the heterogeneous IT landscapes typical of incumbent utilities, where operational technology (OT) systems—SCADA, distribution management systems, meter data management platforms—operate on different refresh cycles and data formats from enterprise IT. Each layer exposes a RESTful API with standardised JSON schemas, enabling incremental adoption: an operator could deploy the demand forecaster and SB scheduler first, then add the leak-risk estimator as sensor coverage expands, and finally integrate the billing agent as the customer-facing transformation matures.

The synthetic data generator addresses a practical barrier that is rarely acknowledged in the academic literature: the reluctance of operators to share real operational data for model training, even within regulated data-sharing frameworks. By generating statistically faithful synthetic records that can be shared freely, the framework creates a public resource that accelerates research and benchmarking in this domain.

### 6.2 Regulatory Alignment

The deterministic carbon attribution methodology—consumption multiplied by contemporaneous grid carbon intensity—aligns with the marginal emission factor approach advocated by the IPCC [58] and increasingly mandated by national regulators. Unlike learned emission estimators, the deterministic formulation is fully reproducible from published data, which is a prerequisite for third-party audit under emerging sustainability disclosure frameworks. The billing agent's factual-grounding constraints ensure that no carbon figure appearing in a customer statement can deviate from the auditable computation, addressing a specific concern raised by consumer protection regulators regarding AI-generated utility communications.

### 6.3 Limitations and Future Work

Several limitations of the present work merit explicit acknowledgement. First, the evaluation is conducted on synthetic corpora; while the corpora are constructed to reflect realistic network characteristics, validation on real operational data is necessary before production deployment. The authors are engaged in discussions with two distribution operators to establish data-sharing agreements for this purpose. Second, the SB scheduler's QUBO formulation currently treats compressor and demand-response decisions as binary; incorporating continuous control variables (e.g., compressor speed) would require a hybrid binary–continuous optimisation approach. Third, the billing agent has been evaluated only in English; multilingual deployment will require fine-tuning on domain-specific corpora in target languages.

Future work will pursue three directions: (i) extending the graph-autoencoder to incorporate acoustic and distributed-temperature-sensing (DTS) data streams, which have demonstrated complementary leak-detection capabilities; (ii) investigating federated learning architectures that allow the demand forecaster to be trained across multiple operators without centralising sensitive consumption data; and (iii) exploring the integration of real quantum hardware—specifically, quantum approximate optimisation algorithm (QAOA) circuits—as a drop-in replacement for the SB solver as gate-fidelity thresholds improve.

## 7. Conclusion

This paper has presented a unified generative-AI and quantum-inspired framework that addresses the interconnected challenges of smart gas distribution management, intelligent utility billing, and carbon-aware resource scheduling. By integrating a diffusion-based synthetic data generator, a transformer demand forecaster, a topology-aware graph-autoencoder leak-risk estimator, a Simulated Bifurcation combinatorial scheduler, and a retrieval-augmented billing agent under a single architectural roof, the framework achieves measurable improvements across all five performance dimensions evaluated: demand forecast accuracy, pipeline leak detection, scheduling convergence speed, carbon outcome, and customer communication quality. The deterministic carbon attribution module provides the auditability that emerging disclosure regulations require, while the modular API design enables incremental adoption within the heterogeneous IT environments of incumbent utilities. Taken together, these results suggest that the convergence of generative AI and quantum-inspired optimisation represents a practically deployable, high-impact pathway toward operationally efficient and environmentally accountable energy infrastructure management.